\documentclass[runningheads]{llncs}
\usepackage[T1]{fontenc}
\usepackage{cite}
\usepackage{amsmath,amssymb,amsfonts}
\usepackage{algorithmicx}
\usepackage{algorithm}
\usepackage{float}
\usepackage{graphicx}
\usepackage{textcomp}
\usepackage{xcolor}
\usepackage{colortbl}
\usepackage{hyperref}
\usepackage{algpseudocode}
\algtext*{EndIf}
\algtext*{EndFor}
\usepackage{multirow}
\usepackage{multicol}
\usepackage{booktabs}
\usepackage{color}
\usepackage{caption}
\usepackage{subcaption}
\usepackage[inline]{enumitem}
\usepackage{graphicx,verbatim}
\begin{document}
\title{Learning What is Worth Learning: Active and Sequential Domain Adaptation for Multi-modal Gross Tumor Volume Segmentation}
%
\begin{comment}  %% Removed for anonymized MICCAI 2025 submission
\author{First Author\inst{1}\orcidID{0000-1111-2222-3333} \and
Second Author\inst{2,3}\orcidID{1111-2222-3333-4444} \and
Third Author\inst{3}\orcidID{2222--3333-4444-5555}}
%
\authorrunning{F. Author et al.}
\institute{Princeton University, Princeton NJ 08544, USA \and
Springer Heidelberg, Tiergartenstr. 17, 69121 Heidelberg, Germany
\email{lncs@springer.com}\\
\url{http://www.springer.com/gp/computer-science/lncs} \and
ABC Institute, Rupert-Karls-University Heidelberg, Heidelberg, Germany\\
\email{\{abc,lncs\}@uni-heidelberg.de}}

\end{comment}

\author{Jingyun Yang\thanks{Jingyun Yang and Guoqing Zhang contributed equally to this work.}, Guoqing Zhang$^\star$, Jingge Wang, and Yang Li} 
%index{Yang, Jingyun}
%index{Zhang, Guoqing}
%index{Wang, Jingge}
%index{Li, Yang}

\authorrunning{J. Yang and G. Zhang et al.}
\titlerunning{Learning What is Worth Learning}
\institute{Shenzhen Key Laboratory of Ubiquitous Data Enabling, Tsinghua Shenzhen International Graduate School, Tsinghua University \\
    \email{yangjy20@mails.tsinghua.edu}}
\maketitle              % typeset the header of the contribution

\begin{abstract}
Accurate gross tumor volume segmentation on multi-modal medical data is critical for radiotherapy planning in nasopharyngeal carcinoma and glioblastoma.
Recent advances in deep neural networks have brought promising results in medical image segmentation, leading to an increasing demand for labeled data. 
% Fine-tuning foundation models on specific low-resource medical tasks has become a standard practice.
Since labeling medical images is time-consuming and labor-intensive, active learning has emerged as a solution to reduce annotation costs by selecting the most informative samples to label and adapting high-performance models with as few labeled samples as possible.
Previous active domain adaptation (ADA) methods seek to minimize sample redundancy by selecting samples that are farthest from the source domain.
However, such one-off selection can easily cause negative transfer, and access to source medical data is often limited. 
% due to various regulatory standards and ethical considerations.
Moreover, the query strategy for multi-modal medical data remains unexplored.
In this work, we propose an active and sequential domain adaptation framework for dynamic multi-modal sample selection in ADA.
We derive a query strategy to prioritize labeling and training on the most valuable samples based on their informativeness and representativeness.
% During each query round, we identify the dominant modality of the selected sample and query its label from the oracle.
Empirical validation on diverse gross tumor volume segmentation tasks demonstrates that our method achieves favorable segmentation performance, significantly outperforming state-of-the-art ADA methods.
Code is available at the git repository: 
\href{https://github.com/Hiyoochan/mmActS}{mmActS}.

% Authors must provide keywords and are not allowed to remove this Keyword section.

\end{abstract}

\section{Introduction}
Precise delineation of the Gross Tumor Volume (GTV) plays a pivotal role in ensuring effective radiotherapy for prevalent malignancies such as nasopharyngeal carcinoma, predominantly impacting the head and neck area \cite{wang2024dual}, and glioblastoma, presumably originating from glial cells and posing a severe threat to human health
\cite{menze2014multimodal}.
Magnetic Resonance Imaging (MRI) is widely used for tumor detection due to its high soft tissue contrast and non-invasive nature, 
% By utilizing multiple modalities rather than relying on a single type of MRI images,
while multi-modal MRI data can map various tumor-induced tissue changes \cite{wang2023learnable}.
% For example, FLAIR MRI highlights differences in tissue water relaxation properties, while post-Gadolinium T1 MRI reveals pathological intratumoral uptake of contrast agents.
For example, FLAIR highlights tissue water relaxation differences, while post-Gadolinium T1 reveals intratumoral contrast uptake \cite{bai2023molecular}.

Manual GTV delineation on multi-modal MRI images is time-consuming and subject to inter-observer variability. Recent advances in deep learning have brought promising results in automatic medical image analysis, yielding successful models for various segmentation tasks \cite{cardoso2022monai,wang2024mamba,ma2024segment}.
However, the generalization capability of these models is limited by the large variability in training data and the lack of labeled data \cite{yang2025adapting}.
% due to intricate anatomical structures and wide-range object scales in medical images.
% Another major challenge is the lack of labeled data, as annotating disease-specific medical images is not only time-consuming but also demands specialty-oriented skills, leading to the problem of few-shot
% domain adaptation (FSDA) \cite{motiian2017few}. 
% Although unsupervised domain adaptation methods address the issues without labeled samples, the improvement is limited \cite{wang2024advancing}, compared to FSDA. 
\begin{figure}[t]
%\vspace{-0.2cm}
    \centering
    \includegraphics[width=0.95\linewidth]{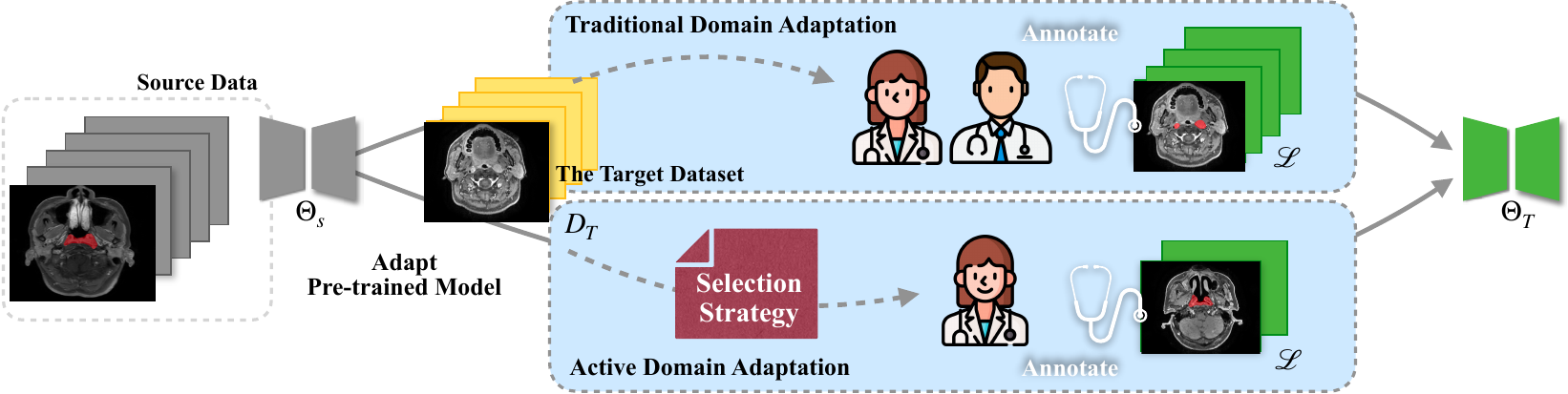}
%\vspace{-0.1cm}
    \caption{Active domain adaptation aims to adapt models pre-trained on adequate source data, which is usually not accessible in medical scenarios, for the desired target task with as few labeled samples as possible.}
    \label{intro}
    %\vspace{-0.3cm}
\end{figure}
One approach to ensuring reliable and robust model adaptation is Active Domain Adaptation (ADA) \cite{wang2024comprehensive}, where the most informative samples are actively selected to be labeled and fine-tune high-performance models with as few labeled samples as possible, as shown in Fig.~\ref{intro}.
Previous ADA methods seek to minimize the sample redundancy either by optimizing for minimal cosine similarity to existing training data \cite{li2024hybrid},
or by combining source feature embeddings as the clustering reference \cite{wang2024dual}.
% , which are not always feasible in real-world medical scenarios with various regulatory standards and ethical considerations.
However, these approaches presuppose access to source data—an assumption that often fails in medical imaging due to regulatory constraints and ethical concerns.
Source-free methods \cite{wang2024comprehensive,tang2023pld} try to select samples with higher epistemic uncertainty based on prediction probabilities with a single forward pass.
Nonetheless, such static one-off selection schemes, which neglect the evolving training dynamics and the inherent domain shift, can easily cause negative transfer. 
% However, these methods haven't fully investigated the characteristics of multi-modal medical images, and such one-off selections can easily cause negative transfer.
Moreover, none of the aforementioned methods fully explore the characteristics of multi-modal medical images, e.g., multi-sequence MRI. 
Works like \cite{cho2022hybrid} utilize transformer layers to integrate features extracted from each modality, while self-attention-based convolution methods enable weighted fusion of multi-modal MRI data \cite{jia2020multi}.
This raises the question: how can multi-modal data be actively combined to better adapt the model?

To address the above issues, we propose an active and sequential domain adaptation framework for advancing gross tumor volume segmentation on multi-modal data in a source-free manner.
To the best of our knowledge, this is the first work to propose a query strategy for multi-modal medical data.
Using a novel dynamic sample selection strategy, we prioritize labeling and training on samples that are worth learning. 
Specifically, in each selection round, we estimate the uncertainty, objective abundance, and density of each sample as indicators of their informativeness and representativeness. 
Taking into account all these factors, we identify the most valuable sample in the current state.
Then, a dominant modality election procedure is introduced to select the modality that exhibits promising performance for annotation, substantially reducing the annotation burden.
By optimizing the use of rare medical resources, both multi-modal data and clinician efforts, 
our method significantly enhances GTV segmentation.
% our method significantly boost the model’s performance.
Extensive experiments on benchmark 3D MRI datasets with various tumor segmentation tasks validate the effectiveness of our method, outperforming all the other state-of-the-art ADA methods. We believe our proposed approach will better leverage rare medical resources, including multi-modal data and clinician expertise, to adapt a model for the desired target task in a fast and scalable way. 
In summary, our main contributions are:
\begin{itemize}
% \vspace{-0.1cm}
    \item \textbf{\textit{An active and sequential domain adaptation framework}}: we propose a novel framework that dynamically selects the most valuable samples to learn,
    enabling an effective adaptation of well-trained source models to target tasks in a source-free manner with as few labeled samples as possible.
    \item \textbf{\textit{An multi-modal sample query strategy}}: we derive an effective query strategy for dynamic multi-modal sample selection and significantly reduce labeling costs, optimizing the use of rare medical resources.
\end{itemize}

%\vspace{-0.2cm}
\section{Methodology}
\begin{figure}[t]
%\vspace{-0.2cm}
    \centering
\includegraphics[width=0.95\linewidth]{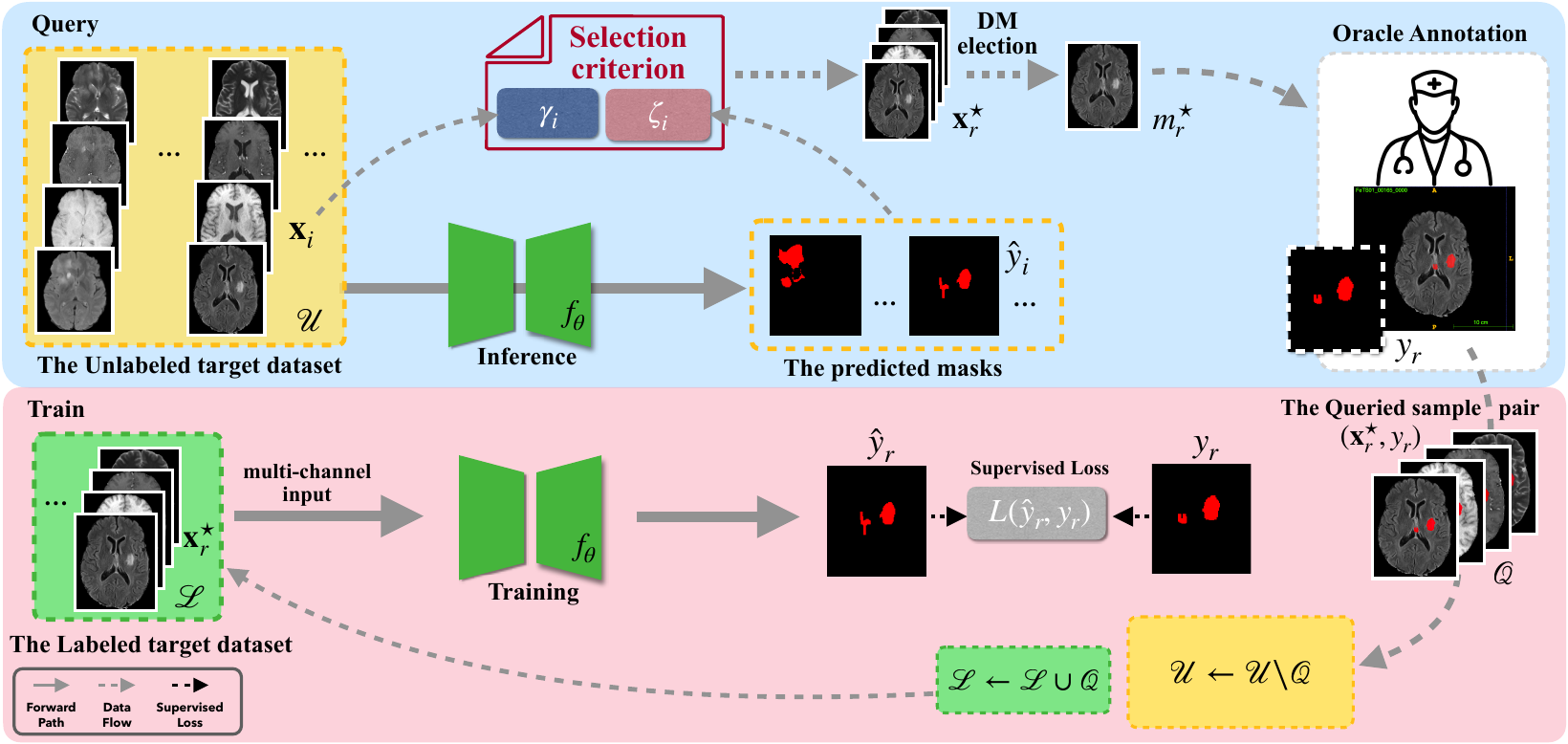}
%\vspace{-0.5cm}
    \caption{Illustration of our active and sequential domain adaptation framework.
    % In the $r$-th query round, the most valuable sample  $\mathbf{x}_r^\star$ is selected based on our criterion. Then $\mathbf{x}_r^\star$ undergoes a DM election procedure, identifying the dominant modality scan $m_r^{(l^\star)}$ for annotation from oracle.
    The blue block shows the $r$-th query round while the pink block represents the target model’s fine-tuning under the supervision of the updated labeled set $\mathcal{L}$.}
    \label{model}
    %\vspace{-0.2cm}
\end{figure}

%\vspace{-0.1cm}
In this section, we present the proposed active and sequential domain adaptation framework in the context of multi-modal medical image segmentation, shown in Fig.~\ref{model}.
First, we clarify the setting of active learning with multi-modal data.
Then we introduce the sequential query strategy and define the selection criterion, with the consideration of the informativeness and representativeness of each sample.
Finally, we describe the target model fine-tuning procedure.
%\vspace{-0.2cm}
\subsection{Problem Definition}
% \subsubsection{ADA problem setting.}

\noindent\textbf{ADA problem setting.} For active learning problems, given a target domain dataset $\mathcal{D}_\mathcal{T}=\{\mathbf{x}_i\}_{i=1}^{n}$,
we select the most valuable samples to label and learn.
% Assuming access to a source model $\Theta_s$ pre-trained on an adequate source domain dataset $\mathcal{D}_\mathcal{S}$, which is not always accessible in real-world medical scenarios with various regulatory standards and ethical considerations,
% In each query round, the agent actively draws samples from a large pool of unlabeled data and requests an oracle (e.g., doctors) for annotation.
Assuming access to a source model $\Theta_s$, pre-trained on adequate source domain data that is not always accessible in medical scenarios,
the goal is to adapt $\Theta_s$ to achieve high performance on $\mathcal{D}_\mathcal{T}$ with as few labeled samples as possible.
%\vspace{-0.3cm}
% \subsubsection{Multi-modal Learning.}
\\\\
\noindent\textbf{Multi-modal Learning.} Let us represent the M-modality sample with $\mathbf{x}_i = \{m_i^{(l)}\}_{l=1}^M$ where $m^{(l)}_i$ is the $l$-th modality scan of sample $\mathbf{x}_i$.
To effectively fuse information from multiple modalities, we train the model in a multi-channel manner. 
Specifically, we stack the different modalities along the channel dimension, forming a multi-channel input fed into the model.
The convolutional layers process all channels jointly, enabling the model to capture both modality-specific and cross-modality features for improved representation learning.
%\vspace{-0.3cm}
\subsection{Query Strategy}
%\vspace{-0.1cm}
Instead of a one-off selection, we sequentially select the most valuable samples until the labeling budget $\mathcal{B}$ is exhausted, as the model evolves and the most informative sample changes at each time step.
We derive a selection criterion to assess each sample in the unlabeled pool $\mathcal{U}\subseteq\mathcal{D}_\mathcal{T}$.
In the $r$-th selection round, the most valuable sample $x^*_r$ is selected and future undergoes a dominant modality (DM) election procedure for all $m^{(l)}_r \in \mathbf{x}^\star_r$ to identify the modality $l^\star$ that exhibits more promising performance than others.
Next we query the label for scan $m^{\star}_r$ from oracle, yielding the labeled pair $ (\mathbf{x}^\star_r,y_r)$.
Then the model is trained on the labeled set $\mathcal{L} = \{(\mathbf{
x}_1^\star,y_1),...,(\mathbf{x}_r^\star,y_r)\}$ from this selection round to the next.
% parameters w.r.t. $\{(x_1^\star,y_1),...,(x_t^\star,y_t)\}$.
%\vspace{-0.3cm}
% \subsubsection{Selection criterion.}
\\\\
\noindent\textbf{Selection criterion.} In domain adaptation tasks, the source model $\Theta_\mathcal{S}$ has already acquired some fundamental knowledge \cite{guan2021domain,yang2025adapting}.
% of medical image segmentation.
% Given the labeling budget, it is essential to learn samples that are worth learning.
% \vspace{-0.5cm}
\begin{algorithm}[!t]
% \vspace{-0.05cm}
  \caption{Active and sequential fine-tuning with Multi-modal data.}
  \label{aif}
  \begin{algorithmic}[1]
    \Require $\mathcal{D}_\mathcal{T}=\{\mathbf{x}_i\}_{i=1}^{n}$: The target dataset
    \Require $f_\theta$: Parameterized model
    %\Require $M$: Best arm selection steps
    \Require $\mathcal{B},\tau$: Labeling budget, query stride
    \State \textbf{Initialize:} $f_{\theta_0}=\Theta_s$ ; $\mathcal{U}\leftarrow\{x_i\}_{i=1}^{n}$ ; $\mathcal{L}\leftarrow\varnothing$ ;
    selection round $r=1$
    % \State \textbf{Initialize:} selection round $r=1$
    \For{$t = 0, 1, \dots , T$}
        \If{$r \leq \mathcal{B} \land t = \frac{r(r-1)}{2}\tau$}
            \State Compute $s(\mathbf{x}),\forall \mathbf{x} \in \mathcal{U}$ according to Equation.\ref{criterion}
            \State Select the most valuable sample $\mathbf{x}^\star_r = \arg\max_{\mathbf{x}\in\mathcal{U}}s(\mathbf{x})$
            \State Identify the dominant modality scan $m^{\star}_r$ according to Equation.\ref{DMEP}
            \State Query the label $y_r$ for $m^{\star}_r$ from oracle, yielding $\mathcal{Q} = (\mathbf{x}_r^\star , y_r) $, $r \leftarrow r + 1 $
            \State Update the labeled and unlabeled sets $\mathcal{L} \leftarrow \mathcal{L} \cup \mathcal{Q} $ ; $ \mathcal{U} \leftarrow \mathcal{U} \setminus \mathcal{Q}$
        \EndIf
        \State \textbf{end if}
        \State Update model parameters w.r.t. $\mathcal{L}$
    \EndFor
    \State \textbf{end for}  \\
    \Return $f_\theta$
      % \vspace{-0.1cm}
  \end{algorithmic}
\end{algorithm}

To achieve good target performance, we estimate the informativeness and representativeness of each sample to capture the complexities and variabilities in the target data.
Given a labeling budget, we prioritize annotating and training on samples that are worth learning, enabling the model to learn robust representation capabilities.

At the time step $t$, we have the current model $f_{\theta_t}$ and obtain the predicted mask $\hat{{y}}_i$ for sample $\mathbf{x}_i$: $\hat{{y}}_i = \arg\max_c(\text{softmax}_c(f_{\theta_t}(\mathbf{x}_i)))$.
% %\vspace{-0.1cm}
% \begin{equation}
% %\vspace{-0.1cm}
%     \hat{{y}}_i = \arg\max_c(\text{softmax}_c(f_{\theta_t}(\mathbf{x}_i)).
% \end{equation}

In the $r$-th selection round,  we quantify the informativeness $\zeta_i$ of sample $\mathbf{x}_i$, by jointly considering its predictive uncertainty $\mu_i$ and the objective abundance, estimated via the total predicted volume:
\begin{equation}
% %\vspace{-0.1cm}
    \zeta_i = \mu_i \sum\nolimits_{v_k\in\hat{{y}}_i} \mathbf{1}(v_{k} = 1),
\end{equation}
% where the summation $\sum_{v_k\in \hat{y}_i}$ denotes the voxel-wise sum in the predicted mask.
where the summation $\sum_{v_k\in \hat{y}_i}$ denotes the number of voxels predicted as foreground in the segmentation mask $\hat{y}_i$.
The uncertainty score $\mu_i$ is computed as the mean voxel-wise entropy across all predictions for sample $\textbf{x}_i$. 
Specifically, for a model output $f_{\theta_t}$ with $C$ classes and $K$ voxels, we define:
% The uncertainty score $\mu_i$ is the mean value of the voxel-wise sum in the uncertainty map $U_i$ with $C$ classes, where $\mathcal{H}(.)$ is the entropy function:
%\vspace{-0.2cm}
\begin{equation}
         \mu_i = \frac{1}{K}\sum_{u_k\in U_i} U_i = \frac{1}{K}\sum_{u_k\in U_i} \sum_{c=1}^{C} \mathcal{H}  (\text{softmax}_c(f_{\theta_t}(\textbf{x}_i)))),
\end{equation}
where $\mathcal{H}(.)$ denotes the entropy function and $U_i$ is the voxel-level uncertainty map.
Inspired by \cite{wang2024advancing}, two components in the measure are interpreted as follows: 
1) \textit{uncertainty} cue, and 2) \textit{concentration} cue. 
The former identifies data that the model cannot predict confidently, while the latter indicates data with a higher concentration of objectives, necessitating oracle annotation for precise model training and refinement. 

% where $\mathcal{H}(.)$ is the entropy function and $C$ represents the classes.
Meanwhile, we quantify the representativeness of $\textbf{x}_i$ based on its density,
%\vspace{-0.2cm}
\begin{equation}
         \gamma_i = \sum\nolimits_{\mathbf{x}_j\in \mathcal{U}, j \neq i}\!e^{-\left(\frac{\omega(\mathbf{x}_i,\mathbf{x}_j)}{\omega_d}\right)^2},
\end{equation}
where $\omega(\mathbf{x}_i,\mathbf{x}_j)$ is the distance between samples $\mathbf{x}_i$ and $\mathbf{x}_j$ measured by the Wasserstein distance \cite{panaretos2019statistical} for its ability to handle shifts in data distributions and $\omega_d$ is the neighborhood distance threshold.
% for its stability and ability to handle shifts in data distributions and $\omega_d$ is the neighborhood distance threshold.
Specifically,
given a pair of samples $(\mathbf{x}_i,\mathbf{x}_j)$ with M modalities, $\omega(\mathbf{x}_i,\mathbf{x}_j)$ is defined as:
%\vspace{-0.1cm}
\begin{equation}
    \omega(\mathbf{x}_i,\mathbf{x}_j) \triangleq  \frac{1}{M}\sum\nolimits^{M}_{l=1}\mathcal{W}(\hat{P}^{(l)}_{i},\hat{P}^{(l)}_{j}),
\end{equation}
% $
%     \omega(x_i,x_j) \triangleq  \frac{1}{M}\sum^{M}_{l=1}\mathcal{W}(\hat{P}^{(l)}_{i},\hat{P}^{(l)}_{j}),
% $
where $(\hat{P}^{(l)}_{i},\hat{P}^{(l)}_{j})$ are distributions of images $(\hat{m}^{(l)}_i,\hat{m}^{(l)}_j)$ after dimension reduction using principal components analysis.
% And the data-pair Wasserstein distance $\mathcal{W}(\hat{P}^{(l)}_{i},\hat{P}^{(l)}_{j})$ is defined as,
And the data-pair Wasserstein distance is defined as:
\begin{equation}
    \mathcal{W}(\hat{P}^{(l)}_{i},\hat{P}^{(l)}_{j}) = \inf _{\gamma \in \Pi\left(\hat{P}^{(l)}_{i},\hat{P}^{(l)}_{j}\right)} \mathbb{E}_{(x, y) \sim \gamma}\|x-y\|.
\end{equation}
% During the active and sequential training, we represent the target dataset by $\mathcal{D}_{\mathcal{T}}=\mathcal{U}\cup\mathcal{L}$ where $\mathcal{U}$ denote the unlabeled samples and $\mathcal{L}$ denotes the queried samples. 
Finally, the selection criterion $s(\mathbf{x}_i)$ for unlabeled target data is written as:
%\vspace{-0.1cm}
\begin{equation}
%\vspace{-0.1cm}
\label{criterion}
   s(\mathbf{x}_i)  = \zeta_i\gamma_i.
 \end{equation}
Accordingly, in the $r$-th selection round, the most valuable sample $\mathbf{x}^\star_r$ is selected,
$\mathbf{x}^\star_r = \arg\max_{\mathbf{x}_i\in\mathcal{U}}s(\mathbf{x}_i)$,
and future undergoes a DM election procedure to identify the dominant modality scan for annotation.
%\vspace{-0.3cm}
% \subsubsection{Label Query.}
\\\\
\noindent\textbf{Label Query.} In the dominant modality election procedure, a validation process is applied:
For the selected sample $\mathbf{x}^\star_r$, the modality that exhibits promising performance is selected to be annotated.
Formally, we have:
%\vspace{-0.1cm}
\begin{equation}
    \begin{aligned}
    \label{DMEP}
     m^{\star}_r = {\arg\max}_{l\in\{1,...,M\}} d(f_{\theta_t}(m_r^{(l)}),\hat{y}_r),
    \end{aligned}
\end{equation}
where the pseudo-label $\hat{y}_r$ is estimated by $f_{\theta_t}(\textbf{x}^\star_r)$, $l$ indexes different modalities, $f_{\theta_t}$ is the current model, and $d(.,.)$ is the function to calculate the dice score.
We query the label for the dominant modality scan $m^{\star}_r$ from oracle (e.g., doctors), yielding the labeled pair $ \mathcal{Q} = (\mathbf{x}^\star_r,y_r)$ and the updated labeled set $\mathcal{L} \leftarrow \mathcal{L} \cup\mathcal{Q}$.
\begin{figure}[!t]
%\vspace{-0.1cm}
    \centering
    %\vspace{-0.1cm}
    \includegraphics[width=\linewidth]{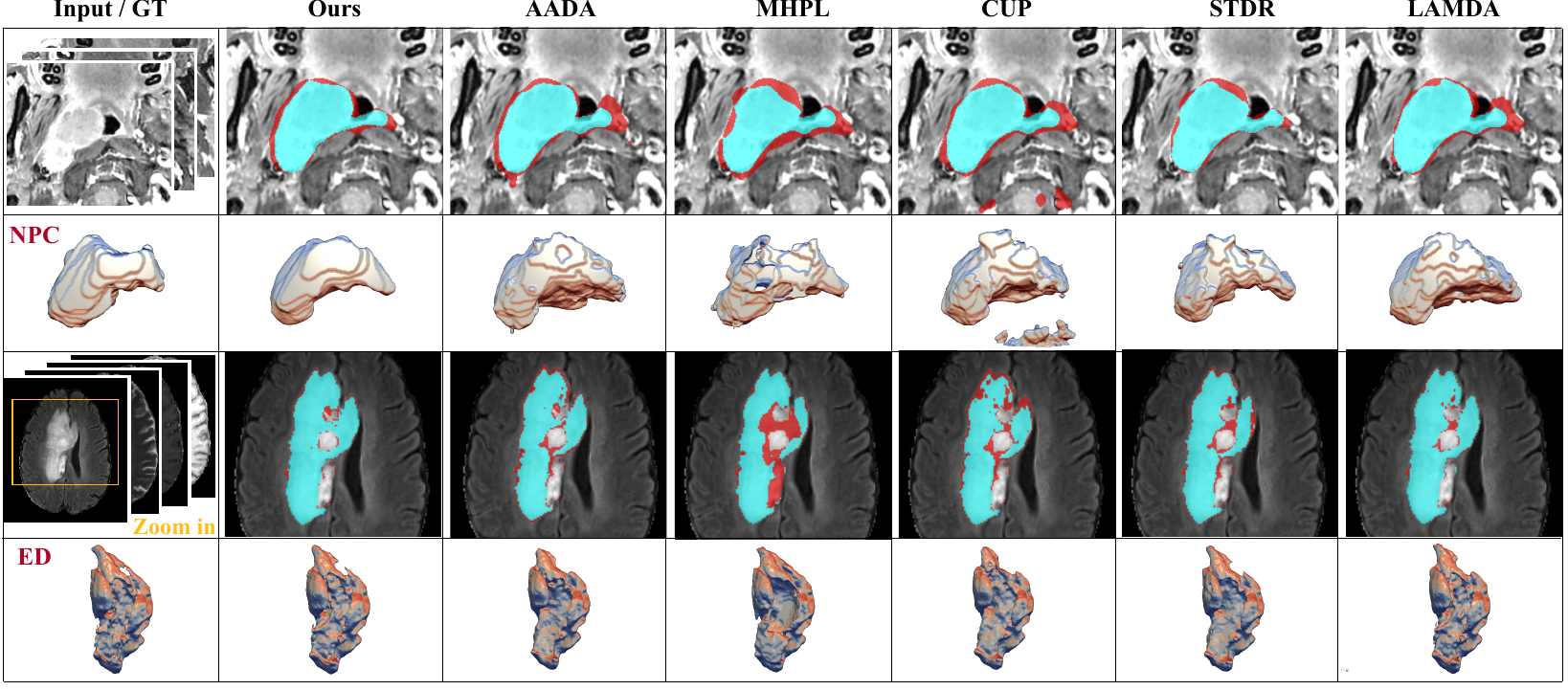}
    %\vspace{-0.6cm}
    \caption{Visualization of SOTA active domain adaptation methods performance. In the first and third rows, the pixels of 2D slices highlighted in red represent \textit{incorrect predictions}, while the second and fourth rows offer 3D comparisons.}
    \label{result}
\end{figure}
\subsection{Active and Sequential Model Training}
%\vspace{-0.1cm}
After the $r$-th oracle annotation round, we train the model using the labeled set $\mathcal{L} = \{(\mathbf{x}^\star_1,y_1),...,(\mathbf{x}^\star_r,y_r)\}$ in a supervised manner.
It is worth noting that if the model initialization is not proper to start the active learning process, it may produce meaningless informativeness estimation for the target domain \cite{wang2024comprehensive}.
% Active learning relies on proper model initialization and accurate sample assessment—a process that, if not executed properly, could lead to suboptimal selection and a decrease in performance gains.
A one-off selection based on such criteria can easily cause negative transfer.
To mitigate this, we sequentially select the most valuable sample to label in each query round, based on the model’s current state, until the labeling budget is reached, i.e., $r = \mathcal{B}$, and fine-tune the model incrementally.
% The final number of labeled samples after the last selection round is far less than the total count of the target samples,
After the final query round, the labeled samples remain significantly fewer than the total target samples.
i.e., $|\mathcal{L}|=\mathcal{B}\ll\mathcal{D}_\mathcal{T}$.
As the extra computational cost for sample assessment is minimal (on the order of seconds), the query strategy adds negligible overhead.
The proposed algorithm is shown in Alg.~\ref{aif}.

\begin{table}[!t]
	\centering
%\begin{minipage}{0.95\linewidth}
\centering
\setlength\tabcolsep{4pt} 
\footnotesize
\caption{Results of active domain adaptation strategies on GTV segmentation 3D MRI datasets BraTS 2022 (ET, NCR, ED) and NPC 2024. 
% The segmentation evaluation metrics are the Dice score and the mean IoU score.
$|\mathcal{L}|$: the number of labeled target samples. Bold number: best score except for the upper bound.}
\label{AIM}
\resizebox{\linewidth}{!}{%
\begin{tabular}{cl|cc|cc|cc|cc}
\toprule
\multicolumn{2}{c}{Method}
 & \multicolumn{2}{c}{ET} & \multicolumn{2}{c}{NCR} & \multicolumn{2}{c}{ED}  & \multicolumn{2}{c}{NPC} \\
\cmidrule(lr){1-2}
\cmidrule(lr){3-4}
\cmidrule(lr){5-6}
\cmidrule(lr){7-8}
\cmidrule(lr){9-10}
\multicolumn{1}{c}{Strategy} & \multicolumn{1}{l}{$|\mathcal{L}|$}\quad & \multicolumn{1}{c}{Dice (\%)} & \multicolumn{1}{c}{mIoU (\%)}
 & \multicolumn{1}{c}{Dice (\%)} & \multicolumn{1}{c}{mIoU (\%)}
 & \multicolumn{1}{c}{Dice (\%)} & \multicolumn{1}{c}{mIoU (\%)}
  & \multicolumn{1}{c}{Dice (\%)} & \multicolumn{1}{c}{mIoU (\%)}\\
\cmidrule(lr){1-10}
\rowcolor{gray!20}Lower bound & 0 & 77.70 & 64.39 & 61.51 & 49.46 & 69.85 & 56.15 & 67.63 & 51.71 \\ 
\rowcolor{gray!20}Upper bound & 80 & 94.10 & 89.20 & 78.94 & 69.72 & 92.25 & 87.37 & 76.12 & 61.85 \\ 
Random Selection & 3 & 78.09 & 67.52 & 60.12 & 51.72 & 73.33 & 59.73 & 66.51 & 51.07\\
AADA (WACV' 20) & 3 & 81.51 & 74.53 & 65.66 & 56.95 & 76.30 & 64.20  & 69.71 & 53.91\\ 
MHPL (CVPR' 23) & 3 & 88.63 & 79.62 & 62.85 & 51.72 & 77.67  & 68.32  & 68.01 & 52.81\\ 
CUP (MICCAI' 24)& 3 & 83.60 & 75.12 & 64.68 & 47.80 & 77.58 & 67.74  & 69.04 & 53.79\\ 
STDR (TMI' 24) & 3 &  88.65 & 81.09 & 62.53 & 50.86 & 78.08 & 68.84  & 71.96 & 56.95\\ 
LAMDA (ECCV'22) & 3 & 89.45 & 80.91  & 66.50 & 49.81 & 78.67 & 68.74 & 71.17 & 56.26 \\
\rowcolor{gray!40}Ours & 3 & \textbf{91.11} & \textbf{84.13} & \textbf{72.30} & \textbf{64.14} & \textbf{82.82} & \textbf{74.10} & \textbf{74.69} & \textbf{59.96}\\ 
 \bottomrule
\end{tabular}
}
%\end{minipage}
\end{table}

\section{Experiments and Results}
\subsection{Datasets and Training Setup}
\label{setting}
Two multi-modal GTV segmentation datasets are used in our work: BraTS 2022 \cite{menze2014multimodal,baid2021rsna,bakas2017advancing} and NPC 2024 \cite{luo2023deep,wang2024dual}.
BraTS 2022 includes 3D MRI volumes across FLAIR, T1, T1c, and T2 modalities, segmenting for enhancing tumor (ET), edema (ED), and necrotic core (NCR).
NPC 2024 is characterized by nasopharyngeal carcinoma (NPC) and manually delineated on each slice of the patient’s T1, T1c, and T2 MRI images.
% For both datasets, we use 100 cases for training under the 1-way scenario using 5-fold cross-validation, and 20 for evaluation.
For both datasets, we use 80 cases for training and 20 for evaluation.
% , with a resolution of $240 \times 240 \times 155$.
We implement all methods on pre-trained nnU-Net \cite{isensee2021nnu} for all experiments, following the pre-training settings of MONAI \cite{cardoso2022monai}. 
% Due to the large image size and memory constraints, we set a batch size of 2.  
% We crop all data to the region of nonzero values in the same size. 
The models are trained using an A800 80GB GPU for a maximum of 600 epochs with a batch size of 5 and an initial learning rate of 0.01, decayed following the poly learning rate policy \cite{chen2017deeplab}.
% We use the Adam optimizer with an initial learning rate of $0.01$ and set it to decrease periodically if the losses do not improve enough.
For the query strategy, the selection stride is set to 40 epochs.
For the label budget, following the common few-shot active learning setting in \cite{ouyang2022self}, all the experiments are conducted under a 1-way 3-shot scenario using 5-fold cross-validation.
We experimented with budgets of labeling 5, 3, 2, and 1 target samples, and found that using 3 samples strikes a good balance, achieving satisfactory performance while keeping annotation costs low.

% For the label budget, we experimented with budgets of labeling 1, 2, 3, and 5 scans, as detailed in our ablation study in Section~\ref{budget}, 
% and found that using a budget of 3 scans yields satisfactory results. Therefore, we report all comparisons with other ADA methods using a label budget of 3.
% To avoid overfitting, we utilize a large variety of data augmentation methods on the fly during training: random rotations, random scaling, and random elastic deformations.
% All MRI images used in experiments are preprocessed via a standard pipeline: registration, skull stripping, and bias field correction.

\subsection{Performance evaluation}
% \vspace{-0.1cm}

\begin{table}[!t]
%\vspace{-0.3cm}
	\centering
%\begin{minipage}{0.95\linewidth}
\centering
\setlength\tabcolsep{4pt} 
\footnotesize
\caption{Ablation study on single- and multi-modal learning in our method.}
\label{modal}
\resizebox{\linewidth}{!}{%
\begin{tabular}{ccc|cc|cc|cc}
\toprule
{\multirow{2}{*}{Modality}}
 & \multicolumn{2}{c}{ET} & \multicolumn{2}{c}{NCR} & \multicolumn{2}{c}{ED}  & \multicolumn{2}{c}{NPC} \\
\cmidrule(lr){2-3}
\cmidrule(lr){4-5}
\cmidrule(lr){6-7}
\cmidrule(lr){8-9}
 & \multicolumn{1}{c}{Dice (\%)} & \multicolumn{1}{c}{mIoU (\%)}
 & \multicolumn{1}{c}{Dice (\%)} & \multicolumn{1}{c}{mIoU (\%)}
 & \multicolumn{1}{c}{Dice (\%)} & \multicolumn{1}{c}{mIoU (\%)}
  & \multicolumn{1}{c}{Dice (\%)} & \multicolumn{1}{c}{mIoU (\%)}\\
\cmidrule(lr){1-9}
 FLAIR & 46.06 & 31.62 & 37.75 & 26.77 & 73.21 & 59.99 & - & - \\ 
 T1 & 13.96 & 8.84 & 35.01 & 25.69 & 51.09 & 38.27 & 64.47 & 50.18 \\ 
 T1c & 87.26 & 79.30 & 64.59 & 55.85 & 59.63 & 47.97 & 69.36 & 53.67 \\ 
 T2 & 22.87 & 14.57 & 39.91 & 28.99  & 63.47 & 51.30 & 66.11 & 51.71 \\ 
\rowcolor{gray!20}Multiple & \textbf{91.11} & \textbf{84.13} & \textbf{72.30} & \textbf{64.14} & \textbf{82.82} & \textbf{74.10} & \textbf{74.69} & \textbf{59.96} \\ 
 \bottomrule
\end{tabular}
%\vspace{-0.6cm}
}
%\end{minipage}
%\vspace{-0.6cm}
\end{table}

To investigate the effectiveness and efficiency of ADA, we consider two baselines:
direct inference without fine-tuning (lower bound) and fine-tuning the model with all samples labeled (upper bound).
Meanwhile, we compare our framework with five state-of-the-art ADA methods, alongside random selection:
1) AADA \cite{su2020active} adversarially adapts the model with importance sampling,
2) MHPL \cite{wang2023mhpl} exploits minimum happy points based on neighbor uncertainty and diversity,
3) STDR \cite{wang2024dual} selects domain-invariant and -specific samples referenced to source domain points,
4) a cascade sampling strategy CUP \cite{wang2024advancing} based on prediction informativeness,
5) a multi-round selection strategy LAMDA \cite{hwang2022combating} with label distribution matching using a density-aware active sampling.
% 6) one-off selection using our selection criterion.
% We keep the label ratio and query stride the same, ensuring a comprehensive analysis under uniform conditions.
We keep the label ratio and query stride the same, ensuring a consistent analysis.
% Meanwhile, to investigate the effectiveness and efficiency of active domain adaptation, we consider two baselines:
% 1) direct inference without fine-tuning (lower bound),
% 2) fine-tuning the model with all samples labeled (upper bound).
% Meanwhile, to investigate the effectiveness of sequential selection,
% we show the results of one-off selection using our selection criterion.
We evaluate the segmentation performance using the Dice score and the mean IoU. 
A quantitative analysis of model adaptation performance on BraTS 2022 and NPC 2024 datasets is detailed in Table.~\ref{AIM} and visualized in Fig.~\ref{result}.
% The BraTS experiment results are the average scores of three target segmentation tasks: ED, ET, and NCR.
The results are averaged over three independent runs with different training data splits to ensure robustness.
Our proposed framework significantly outperforms all state-of-the-art ADA methods on both datasets across anatomical regions.
% achieving an average gain of 9.45\% on BraTS 2022 and 6.81\% on NPC 2024 datasets in Dice score.
Compared to random selection, it achieves an average Dice score gain of \textbf{16.62\%} on BraTS 2022 and \textbf{12.23\%} on NPC 2024.
Furthermore, compared to LAMDA, our method yields an average Dice gain of \textbf{5.28\%} on BraTS 2022 and \textbf{4.95\%} on NPC 2024.
% \subsubsection{Effectiveness of sequential selection.}
\\\\
% \noindent\textbf{Effectiveness of sequential selection.} Active learning relies on proper model initialization and accurate sample assessment; if not executed properly, it could lead to suboptimal selection.
% —a process that, if not executed properly, could lead to suboptimal selection.
% and a decrease in performance gains.
% While LAMDA benefits from multi-round sampling, it requires source data for label matching and joint training, which may not be feasible in medical scenes.
% which is not always feasible in medical scenarios.
% LAMDA performs relatively well with multi-round sampling while benefiting from label distribution matching and joint training with source data, which is not always feasible in medical scenarios.
% We also conduct one-off selection experiments using our criterion, achieving an average Dice of 0.8072 on BraTS and 0.7003 on NPC, compared to 0.8208 (BraTS) and 0.7469 (NPC) with AIM, and 0.7642 (BraTS) and 0.7117 (NPC) with STDR. 
\noindent\textbf{Effectiveness of sequential selection.} Adequate source data for label matching and joint training is usually not available in medical scenes. To further investigate the effectiveness of sequential selection, we conducted one-off selection experiments.
For the NPC 2024 dataset, one-off selection with our criterion achieves a Dice score of 0.7103, compared to 0.7117 with LAMDA, 0.7196 with STDR, and \textbf{0.7469} with our sequential selection.
For the BraTS 2022 dataset, our one-off selection yields an average Dice score of 0.8072, outperforming LAMDA (0.7821) and STDR (0.7642), while our sequential selection achieves the highest score of \textbf{0.8208}. 
% These BraTS results are averaged over three target segmentation tasks: ED, ET, and NCR.
These results highlight the superiority of our sequential selection scheme without relying on any source domain data while achieving SOTA performance.
% \vspace{-0.1cm}
% \subsubsection{Effectiveness of multi-modal learning.}
\\\\
\noindent\textbf{Effectiveness of multi-modal learning.} The results in Table.~\ref{modal} show that multi-modal learning can indeed enhance tumor segmentation.
% Since multi-sequence data is easily collected, clinicians typically provide scans with multiple modalities.
% Rather than randomly choosing a single modality, especially given the performance variations observed in ET results, leveraging multiple modalities is a better way to optimize the use of rare medical resources.
Clinicians often provide scans with multiple modalities since multi-sequence data is easily collected, making it more efficient to use all available data rather than randomly choosing a single modality, especially given the performance variations observed in ET results.
% We believe our multi-modal data query strategy can drive meaningful advancements in medical image processing.
% \vspace{-0.4cm}
% \subsubsection{Ablation study on label budget.}
% \label{budget}
% We investigate the trade-off between labeling costs and segmentation performance, with ablation results shown in Table~\ref{budgets}. The 3-budget setting achieves the best balance, offering strong performance with minimal labeling costs.
% Notably, while the source model performs well in the ET target domain, our method further improves performance with just 3 labeled samples, achieving a Dice score of 0.9111, comparable to the fully supervised model trained with 80 labeled samples, significantly reducing labeling costs.

% \vspace{-0.2cm}

% \vspace{-0.2cm}
\section{Conclusion and Future Work}
We propose a novel source-free active and sequential domain adaptation framework for advancing GTV segmentation on multi-modal medical data.
% With our desiderata in mind, the proposed method achieves:
% 1) effective and efficient domain adaptation, delivering strong performance on target tumor segmentation tasks while reducing the annotation burden,
% 2) dynamic prioritized labeling and learning with a query strategy tailored for multi-modal data,
% 3) no requirement for access to the source domain or a substantial amount of labeled target data.
Experiments on two benchmark medical datasets demonstrate that the proposed method achieves state-of-the-art performance in ADA problems within the realm of medical image processing. 
%We believe our proposed approach will better leverage rare medical resources, including multi-modal data and clinician expertise, to adapt a model for the desired target task in a fast and scalable way. 
One limitation of our method lies in the informativeness criterion that is adapted from \cite{wang2024advancing}, originally proposed for vessel segmentation tasks.
While it effectively identifies salient regions, it tends to bias the selection toward larger tumors. 
This can be problematic in datasets with varying tumor sizes, leading to suboptimal performance on early-stage or small-sized tumors.
Moreover, our current experiments are limited to U-Net-based architectures.
For future research, we plan to incorporate more advanced foundation models such as Med-SAM\cite{ma2024segment} and explore robust sampling strategies to improve segmentation performance on small but significant lesions and enhance the overall clinical utility of our approach.

\begin{credits}
\subsubsection{\ackname} This work is supported in part by the Natural Science Foundation of China (Grant 62371270).

\subsubsection{\discintname}
The authors have no conflicts of interest to declare that are relevant to the content of this article.
\end{credits}

\bibliographystyle{splncs04}
\bibliography{main}





% LNCS Homepage, \url{http://www.springer.com/lncs}, last accessed 2023/10/25
% \end{thebibliography}
\end{document}